\documentclass[10pt,twocolumn,letterpaper]{article}

\usepackage{cvpr}              

\usepackage{bm}
\usepackage{multirow}
\usepackage{bbding}
\usepackage{wrapfig}

\usepackage[accsupp]{axessibility}
\usepackage[dvipsnames]{xcolor}

\newcommand{\cparagraph}[1]{{\vspace{+1mm}\noindent\textbf{#1}\quad}}

\definecolor{cvprblue}{rgb}{0.21,0.49,0.74}
\usepackage[pagebackref,breaklinks,colorlinks,allcolors=cvprblue]{hyperref}

\def\paperID{2281} 
\def\confName{CVPR}
\def\confYear{2025}

\title{MIDI: Multi-Instance Diffusion for Single Image to 3D Scene Generation}

\author{
    {Zehuan Huang\textsuperscript{1} \quad
    Yuan-Chen Guo\textsuperscript{2}\footnotemark[2] \quad
    Xingqiao An\textsuperscript{3} \quad
    Yunhan Yang\textsuperscript{4} \quad
    Yangguang Li\textsuperscript{2} \quad
    Zi-Xin Zou\textsuperscript{2}}\\
    \vspace{0.5em}{Ding Liang\textsuperscript{2} \quad
    Xihui Liu\textsuperscript{4} \quad
    Yan-Pei Cao\textsuperscript{2 \Envelope} \quad
    Lu Sheng\textsuperscript{1 \Envelope}} \\
    {\textsuperscript{1}Beihang University \quad
    \textsuperscript{2}VAST \quad
    \textsuperscript{3}Tsinghua University \quad
    \textsuperscript{4}The University of Hong Kong} \\
    {
    Project page: \url{https://huanngzh.github.io/MIDI-Page/}
    }
}

\begin{document}

\twocolumn[
    \maketitle
    \vspace{-2.8em}
    \begin{center}
\centering
\includegraphics[width=\textwidth]{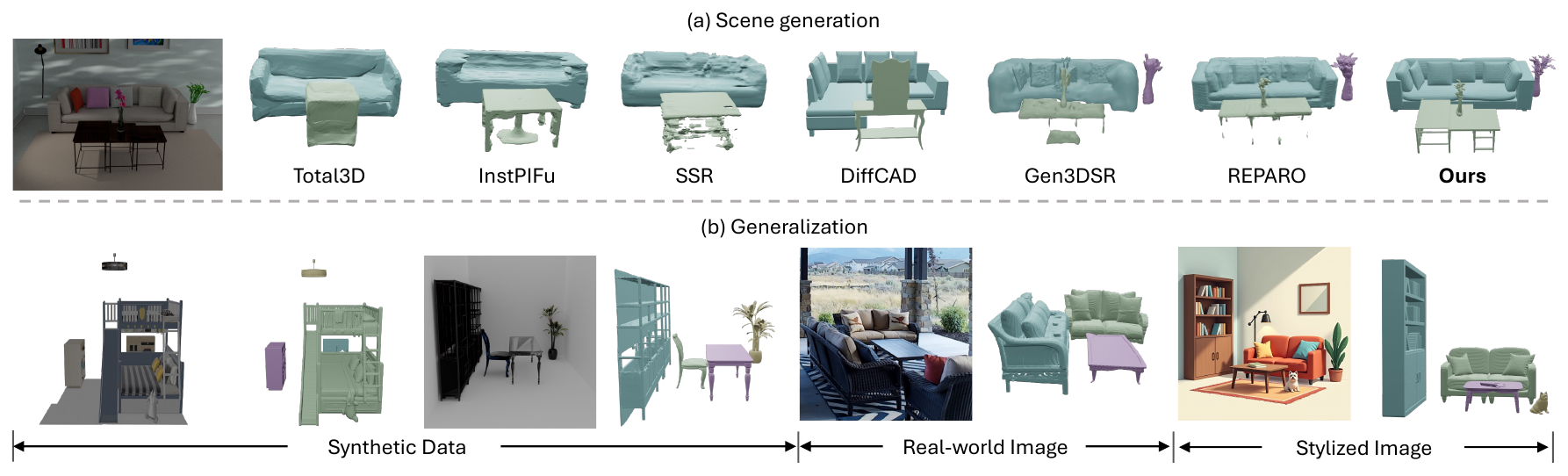}
\captionof{figure}{MIDI generates compositional 3D scenes from a single image by extending pre-trained image-to-3D object generation models to multi-instance diffusion models, incorporating a novel multi-instance attention mechanism that captures inter-object interactions. (a) shows our generated scenes compared with those reconstructed by existing methods. (b) presents our generated results on synthetic data, real-world images, and stylized images.}
\label{fig:teaser}
\end{center}
    \bigbreak
]

\renewcommand{\thefootnote}{\fnsymbol{footnote}}
\footnotetext[2]{Project lead; \textsuperscript{\Envelope} corresponding author}

\begin{abstract}
This paper introduces MIDI, a novel paradigm for compositional 3D scene generation from a single image.
Unlike existing methods that rely on reconstruction or retrieval techniques or recent approaches that employ multi-stage object-by-object generation, MIDI extends pre-trained image-to-3D object generation models to multi-instance diffusion models, enabling the simultaneous generation of multiple 3D instances with accurate spatial relationships and high generalizability.
At its core, MIDI incorporates a novel multi-instance attention mechanism, that effectively captures inter-object interactions and spatial coherence directly within the generation process, without the need for complex multi-step processes.
The method utilizes partial object images and global scene context as inputs, directly modeling object completion during 3D generation.
During training, we effectively supervise the interactions between 3D instances using a limited amount of scene-level data, while incorporating single-object data for regularization, thereby maintaining the pre-trained generalization ability.
MIDI demonstrates state-of-the-art performance in image-to-scene generation, validated through evaluations on synthetic data, real-world scene data, and stylized scene images generated by text-to-image diffusion models.
\end{abstract}

\section{Introduction}
\label{sec:intro}

Generating compositional 3D scenes from a single image is challenging due to the limited spatial clues captured from a partial point of view.
In fact, accurately inferring the 3D geometry of each instance and the spatial relationships of multiple instances within a scene, requires extensive prior knowledge of the 3D visual world.

Existing methods can be broadly categorized into two classes, according to how the prior knowledge is processed. 
The former class~\cite{nie2020total3d,zhang2021im3d,paschalidou2021atiss,dahnert2021panorecon,liu2022instpifu,gkioxari2022usl,zhang2023uni3d,chu2023buol,chen2024ssr} encodes 3D geometry by neural networks that are trained from scene-level 3D datasets, and then inferences the geometry in a new image with a feed-forward pass.
Due to the scarcity of supervised data, these methods often suffer from poor reconstruction quality in unseen scenarios.
The other class~\cite{gumeli2022roca,izadinia2017im2cad,kuo2020mask2cad,kuo2021patch2cad,langer2022sparc,gao2024diffcad} stores 3D models in a database, then retrieve and assemble similar 3D models to match the input image.
However, the limited geometric clues from a single image make it difficult to precisely identify and arrange the correct models.
Moreover, since it is impractical for a 3D database to contain every possible model that exactly corresponds to the input image, the retrieved models can only approximately align with the objects, leading to inconsistencies.
Therefore, methods in both classes lack accuracy and sufficient out-of-domain generalizability, in terms of novel object shapes and unseen scene layouts.

\begin{figure}
    \centering
    \includegraphics[width=\linewidth]{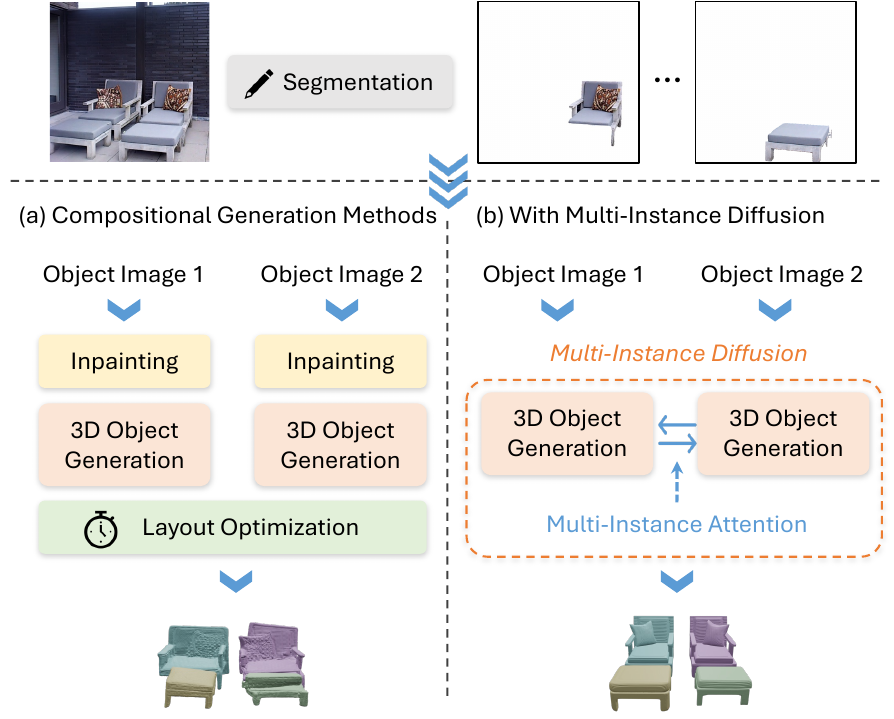}
    \caption{Comparison between our scene generation pipeline with multi-instance diffusion and existing compositional generation methods.}
    \label{fig:pipeline_diff}
\end{figure}

Recent image-to-3D object generation models~\cite{jun2023shape,liu2024one2345,liu2023syncdreamer,long2024wonder3d,hong2023lrm,tang2025lgm,huang2024epidiff,zhang2024clay,wu2024unique3d,li2024craftsman,wen2024ouroboros3d,xu2024instantmesh,voleti2025sv3d,wang2024crm,liu2024one2345++,zhao2024michelangelo,wu2024direct3d,tochilkin2024triposr,guo2023threestudio,yu2024hifi}, with powerful 3d prior and generalization capabilities, can generate high-quality geometry from a single object image.
Building upon these pre-trained models, a common approach for scene generation involves using them as tools within a multi-step compositional generation process, which includes segmenting the scene image, completing individual object images, generating each object, and optimizing their spatial relationships~\cite{zhou2024deepprior,chen2024comboverse,han2024reparo}, as shown in \cref{fig:pipeline_diff}.
While these methods leverage the priors of 3D object generation models, the generation process is inherently lengthy and prone to error accumulation -- errors in intermediate steps can significantly distort the final result.
Moreover, the optimization of spatial relationships cannot directly optimize 3D objects generated one by one by the previous stage that lacks global scene context, leading to misalignments between the generated instances and the overall scene.
Therefore, if inter-object spatial relationships can be modeled directly within the 3D generation model, it is possible to construct an end-to-end pipeline that addresses these issues by generating all instances simultaneously with coherent spatial arrangements.

We propose MIDI, which extends pre-trained 3D object generation models to multi-instance diffusion models, establishing a new paradigm for compositional 3D scene generation.
Our approach enables the simultaneous creation of multiple 3D instances with accurate spatial relationships from a single scene image, moving beyond independent object generation to a holistic understanding of the scene.
Building upon large-scale pre-trained image-to-3D object generation models~\cite{zhang2024clay,wu2024direct3d,li2024craftsman,zhao2024michelangelo}, MIDI employs a novel multi-instance attention mechanism that effectively captures complex inter-object interactions and spatial coherence directly within the generation process, eliminating the need for complex multi-step procedures.
This advanced design allows for the direct generation of cohesive 3D scenes, significantly enhancing both efficiency and accuracy.
Due to the universal nature of spatial relationships between objects, we effectively supervise the interactions between 3D instances using a limited amount of scene-level datasets~\cite{fu20213dfront,fu20213dfuture} during training.
Additionally, we incorporate single-object data for regularization, thereby maintaining the generalization ability of the pre-trained model.

To validate the effectiveness of our proposed paradigm, we conduct experiments on synthetic datasets~\cite{fu20213dfront,fu20213dfuture}, real-world scenes~\cite{dai2017scannet,straub2019replica}, and various stylized scene images generated by text-to-image diffusion models~\cite{rombach2022ldm,podell2023sdxl}.
Results demonstrate that MIDI significantly advances the field of 3D scene generation by effectively modeling inter-object interactions through our multi-instance attention mechanism in the pre-trained 3D generation model.
MIDI produces high-quality 3D scenes with accurate geometries and spatial layouts, while exhibiting strong generalization capabilities.
In summary, our main contributions are as follows:
\begin{itemize}
    \item We establish a new paradigm for compositional 3D scene generation by proposing a multi-instance diffusion model, which extends pre-trained image-to-3D object generation models to generate spatially correlated 3D instances.
    \item We introduce a novel multi-instance attention mechanism that effectively models cross-instance interactions, ensuring the coherence and accurate spatial relationships.
    \item Experiments demonstrate MIDI achieves state-of-the-art performance, significantly improving the generation of 3D scenes by accurately capturing inter-object relationships and providing better alignment with the input.
\end{itemize}

\begin{figure*}
    \centering
    \includegraphics[width=\textwidth]{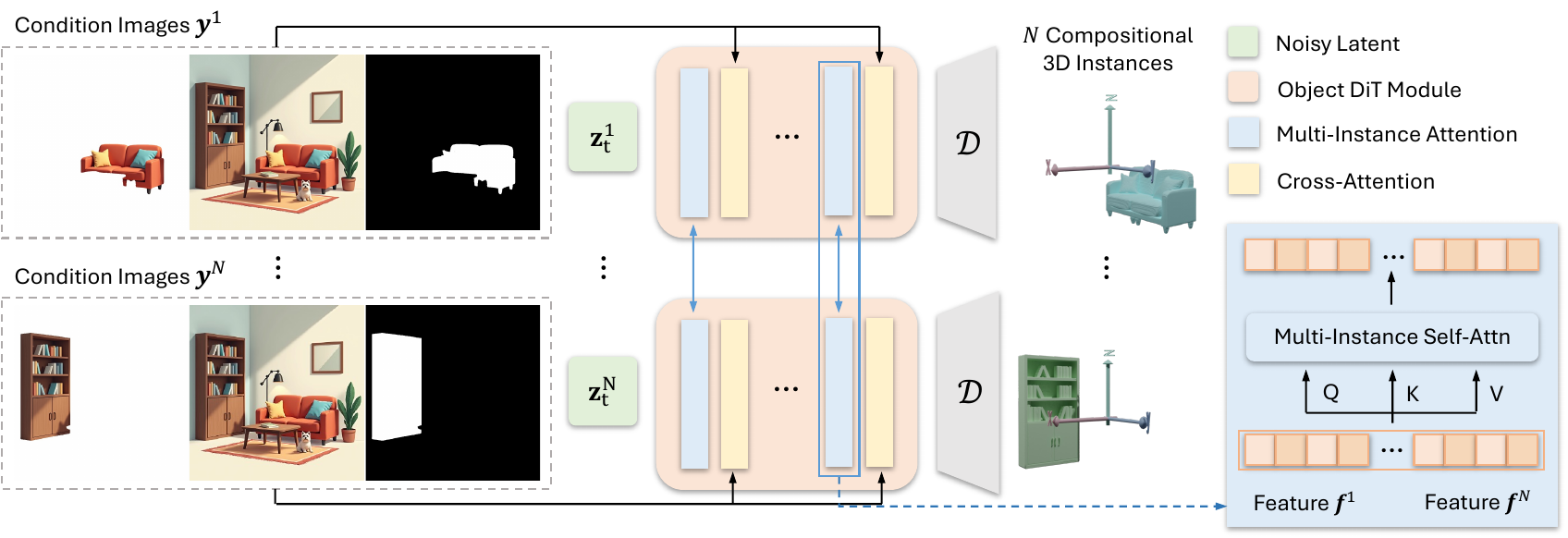}
    \caption{Method overview. Based on 3D object generation models, MIDI denoises the latent representations of multiple 3D instances simultaneously using a weight-shared DiT module. The multi-instance attention layers are introduced to learn cross-instance interaction and enable global awareness, while cross-attention layers integrate the information of object images and global scene context.}
    \label{fig:overview}
\end{figure*}

\section{Related Work}
\label{sec:related_work}

\subsection{Scene Reconstruction from a Single Image}

Recovering the 3D structure of a scene from a single image is a fundamental challenge in computer vision.
Existing methods can be categorized into feed-forward reconstruction methods~\cite{nie2020total3d,zhang2021im3d,paschalidou2021atiss,dahnert2021panorecon,liu2022instpifu,gkioxari2022usl,zhang2023uni3d,chu2023buol,chen2024ssr}, retrieval-based methods~\cite{gumeli2022roca,izadinia2017im2cad,kuo2020mask2cad,kuo2021patch2cad,langer2022sparc,gao2024diffcad}, and recent compositional generation approaches~\cite{zhou2024deepprior,chen2024comboverse,han2024reparo,dogaru2024gen3dsr,tang2024diffuscene}.

\cparagraph{Feed-forward reconstruction methods.}
Feed-forward reconstruction methods~\cite{nie2020total3d,zhang2021im3d,paschalidou2021atiss,dahnert2021panorecon,liu2022instpifu,gkioxari2022usl,zhang2023uni3d,chu2023buol,chen2024ssr} leverage 3D spatial knowledge and use 3D supervision to train end-to-end regression systems.
They typically employ encoder-decoder architectures to predict scene properties such as geometry and instance labels from a single image.
While jointly predicting scene layout and object poses ensures intrinsic correctness, these methods often suffer from limited reconstruction quality due to the scarcity of supervised 3D scene data and struggle to generalize to out-of-distribution images.

\cparagraph{Retrieval-based methods.}
Retrieval-based methods~\cite{gumeli2022roca,izadinia2017im2cad,kuo2020mask2cad,kuo2021patch2cad,langer2022sparc,gao2024diffcad} reconstruct scenes by retrieving and aligning 3D models from a database based on the input image.
For example, DiffCAD~\cite{gao2024diffcad} trains diffusion models~\cite{ho2020ddpm,song2020ddim,song2020scorebased,sohl2015deepun} under synthetic data supervision, to model distributions of CAD object shapes, poses, and scales, which facilitates CAD model retrieval and alignment to the image input.
Although these methods can produce detailed objects by leveraging existing 3D assets, they heavily depend on database diversity and often face retrieval errors due to insufficient information from single images, leading to misalignments.

\cparagraph{Compositional generation methods.}
Recent compositional generation methods~\cite{zhou2024deepprior,chen2024comboverse,han2024reparo,dogaru2024gen3dsr} utilize large-scale perceptual and generative models in both image~\cite{kirillov2023sam,liu2023groundingdino,ren2024groundedsam,ramesh2022dalle2,nichol2022glide,saharia2022imagen,rombach2022ldm,podell2023sdxl} and 3D object~\cite{jun2023shape,eftekhar2021omnidata,zhang2024clay} domains to improve scene reconstruction.
These methods typically involve a multi-stage pipeline, including image segmentation~\cite{ren2024groundedsam}, object completion~\cite{rombach2022ldm}, per-object generation~\cite{jun2023shape,zhang2024clay}, and layout optimization.
While they enhance generalization capabilities by leveraging pre-trained models, their complex pipelines are susceptible to error accumulation, and the lack of global scene context during per-object processing can lead to misaligned results.
Our work addresses these issues by leveraging a pre-trained image-to-3D object generation model to simultaneously generate multiple 3D instances with interrelated relationships, improving robustness and maintaining strong generalization.

\subsection{3D Object Generation from a Single Image}
Advancements in diffusion models~\cite{ho2020ddpm,song2020ddim} and large-scale datasets~\cite{deitke2023objaverse,deitke2024objaversexl} have propelled progress in 3D generation~\cite{liu2024one2345,liu2023syncdreamer,long2024wonder3d,hong2023lrm,tang2025lgm,huang2024epidiff,zhang2024clay,wu2024unique3d,li2024craftsman,wen2024ouroboros3d,xu2024instantmesh,voleti2025sv3d,wang2024crm,liu2024one2345++,wu2024direct3d,zhao2024michelangelo,roessle2024l3dg,wu2024blockfusion,meng2024lt3sd,liu2024part123,dong2025tela}.
Several image-to-3D object generation methods~\cite{liu2023syncdreamer,long2024wonder3d,tang2025lgm,wen2024ouroboros3d,xu2024instantmesh,wang2024crm,voleti2025sv3d,huang2024mvadapter} adopt a two-stage pipeline that involves generating multi-view images and then reconstructing 3D objects.
They fine-tune pre-trained image~\cite{rombach2022ldm,podell2023sdxl} or video~\cite{blattmann2023svd} diffusion models to produce multi-view images and employ large reconstruction models~\cite{hong2023lrm,tang2025lgm,xu2024grm,zou2024tgs} or optimization-based methods~\cite{wang2021neus} to recover geometries.
Another group of work~\cite{zhang2024clay,li2024craftsman,wu2024direct3d,zhao2024michelangelo,li2025triposg} focuses on generating 3D native geometry by training large-scale generative models, which typically comprise a variational autoencoder~\cite{kingma2013vae} and a latent diffusion transformer (DiT)~\cite{peebles2023dit}.
These models produce high-quality geometries with strong generalization due to training on diverse datasets.
Building upon these advancements, we fine-tunes such an object geometry generator to create compositional instances while retaining generalization ability.

\section{Preliminary: 3D Object Generation Models}
\label{sec:preliminary}

Large-scale 3D object generation models~\cite{zhao2024michelangelo,zhang2024clay,wu2024direct3d,li2024craftsman,li2025triposg} are the foundation of our approach.
These models often comprise three main components:
1) a transformer-based variational autoencoder (VAE)~\cite{kingma2013vae} with an encoder $\mathcal{E}$ and a decoder $\mathcal{D}$, which compress 3D geometric representations into a low-dimensional latent space,
and 2) a denoising transformer network $\epsilon_{\theta}$, trained on the compressed latent space to transform noise $\bm{\epsilon}\sim\mathcal{N}(0,I)$ into the original 3D data distribution $\mathbf{z}_{0}$,
and 3) a group of condition encoders, such as CLIP~\cite{radford2021clip} and DINO encoders~\cite{oquab2023dinov2} for encoding text or image conditions, which are then passed to the denoising network by cross-attention mechanism.

At inference time, the denoising process generates samples in the latent space, and the decoder $\mathcal{D}$ produces geometric representations like SDF values or tri-plane features, which can be converted into a 3D mesh by applying marching cubes~\cite{lorensen1998mc} or using an additional mapping network.

\section{MIDI: Multi-Instance 3D Generation}
\label{sec:MIDI}

MIDI lifts 3D object generation to compositional 3D instance generation, enabling the creation of 3D scenes with accurate spatial relationships from a single image.
Specifically, given a scene image, our objective is to generate spatially correlated 3D latent tokens $\{ \mathbf{z}_{0}^{i} \}_{i=1}^{N}$ corresponding to the $N$ instances present in the image.
These latent tokens can be decoded and directly combined to obtain high-quality 3D scenes.

In this section, \cref{sec:multi_instance} first introduces the overall framework of multi-instance diffusion models, detailing how it generalizes single-object diffusion models to handle multiple interacting instances.
\cref{sec:attention} then elaborates on the multi-instance attention mechanism that models cross-instance relationships in 3D space.
Finally, \cref{sec:training} presents the training procedure of MIDI.

\subsection{Multi-Instance Diffusion Models}
\label{sec:multi_instance}

As demonstrated by \cref{fig:overview}, our multi-instance diffusion models extend the original DiT modules of 3D object generation models in three aspects:
1) the latent representations of multiple 3D instances are denoised simultaneously (\ie in parallel) using a shared denoising network, 2) a novel multi-instance attention mechanism is introduced into the DiT modules to learn cross-instance interaction and enable global awareness, and 3) a simple yet effective method for encoding image inputs, including local object images and global scene context.

\cparagraph{Overview of framework.}
Our multi-instance diffusion model builds upon existing 3D object diffusion models by extending them to denoise the 3D representations of multiple instances simultaneously.
Specifically, we retain the VAE of the base model to compress the 3D geometric representations of multiple instances into low-dimensional latent features $\{ \mathbf{z}_{0}^{i} \}_{i=1}^{N}$.
We extend the denoising network $\epsilon_{\theta}$ to condition on the global scene image $\bm{c}_{g}$, the RGB images of the $N$ local objects $\{ \bm{c}_{l}^{i} \}_{i=1}^{N}$, and their corresponding masks $\{ \bm{m}_{l}^{i} \}_{i=1}^{N}$.
The denoising network learns to transform noise $\{\bm{\epsilon}^{i}\sim\mathcal{N}(0,I)\}_{i=1}^{N}$ into the 3D data distribution, effectively capturing the spatial configurations of the instances.

\cparagraph{Cross-instance interaction.}
Compositional 3D instance generation requires that the generated multiple instances exhibit interactive relationships in 3D space.
To achieve this, we introduce a multi-instance attention mechanism within the denoising process, which models cross-instance interactions in the latent feature space during denoising.
The integration of this mechanism transforms the generation of multiple objects from independent processes into a synchronous interactive process, enhancing global scene coherence and ensuring that the spatial relationships among objects are accurately represented.

\cparagraph{Image conditioning.}
To encode all the image conditions, we propose a simple yet effective method, involving 1) the encoding of both global scene information and local instance details and locations with a ViT-based image encoder $\tau_{\theta}$~\cite{oquab2023dinov2}, and 2) integrating the image embeddings using cross-attention layers.
Specifically, for each instance $z^{i}$, we concatenate its RGB image $\bm{c}_{l}^{i}$, mask $\bm{m}_{l}^{i}$, and the global scene image $\bm{c}_{g}$ along the channel dimension, resulting in a composite representation $\mathbf{y} \in \mathbb{R}^{h\times w\times c}$, where $c=7$. The composite image is then passed into a ViT-based encoder with extended input channels to extract a sequence of image features.
Finally, we use a cross-attention mechanism in the transformer-based denoising network to integrate the conditioning image features.

\begin{figure}
    \centering
    \includegraphics[width=0.95\linewidth]{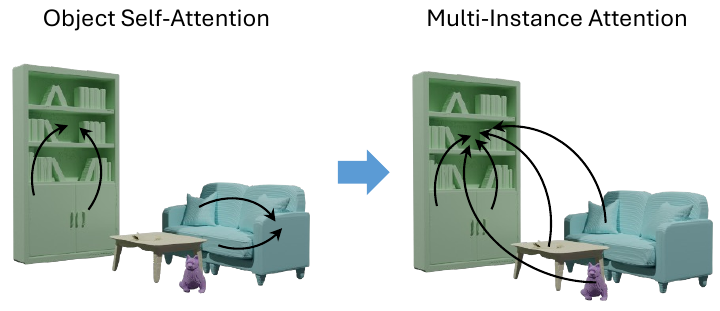}
    \caption{Multi-instance attention. We extend the original object self-attention, where tokens of each object query only themselves, to multi-instance attention, where tokens of each instance query all tokens from all instances in the scene.}
    \label{fig:multi_instance_attn}
\end{figure}

\subsection{Multi-Instance Attention}
\label{sec:attention}

We now introduce the multi-instance attention mechanism, which is the key of MIDI to enforce spatial relationship across multiple 3D instances.
This mechanism extends original object self-attention layers by connecting different instances within the attention computation (See \cref{fig:overview}).

Specifically, we transform the $K$ original object self-attention layers into multi-instance attention layers by integrating the features of all instances $\{ \bm{f}^{i} \}_{i=1}^{N}$ into the attention process, formulated as:
\begin{equation}
\bm{f}^{i}_{\text{out}} = \text{Attention}\left( \bm{f}^{i}, \{ \bm{f}^{j} \}_{j=1}^{N} \right),
\end{equation}
where $\bm{f}^{i}$ is the feature of instance $i$, and $\text{Attention}(\cdot)$ denotes the attention function that allows each instance to attend to the features of all instances in the scene, including itself.
Therefore, as illustrated in \cref{fig:multi_instance_attn}, each token within a particular instance now queries information from tokens of all instances in the scene.
This enables the attention mechanism to effectively model cross-instance interactions by considering the collective set of tokens, thereby capturing inter-object relationships and spatial dependencies.

\subsection{Training}
\label{sec:training}

To train MIDI, we extend the loss of our base model, which utilizes the rectified flow~\cite{liu2022rflow} architecture, from single-object to multi-instance.
For each training step, we sample a shared noise level $t$ from $0$ to $1$ for all the instances $\{ \mathbf{z}^{i} \}_{i=1}^{N}$, perturbing them along a simple linear trajectory:
\begin{equation}
    \{ \mathbf{z}_{t}^{i} \}_{i=1}^{N} = t \{ \mathbf{z}_{0}^{i} \}_{i=1}^{N} + (1-t) \{ \bm{\epsilon}^{i} \}_{i=1}^{N},
\end{equation}
where $\bm{\epsilon}^{i}\sim\mathcal{N}(0,I)$. Then we employ the following loss function to fine-tune the denoising network $\epsilon_{\theta}$ and the image encoder $\tau_{\theta}$:
\begin{equation}
     \mathbb{E}_{\{\mathbf{z}^{i}\}_{i=1}^{N}, \mathbf{y}, \{ \bm{\epsilon}^{i} \}_{i=1}^{N}, t} \Big[
     \sum_{i=1}^{N}
     \lVert \mathbf{z}_{0}^{i} - \bm{\epsilon}^{i} - \epsilon_{\theta}(\mathbf{z}_{t}^{i}, t, \tau_{\theta}(\mathbf{y})) \rVert_{2}^{2}
     \Big].
\end{equation}

Since our training dataset is much smaller than the pre-training dataset of single-object 3D generation models, we incorporate additional 3D object datasets for training to retain the original generalization capability.
In practice, with a $30\%$ chance, we train the multi-instance diffusion model as a simple image-to-3D object generation model on a subset of Objaverse dataset~\cite{deitke2023objaverse} by turning off the multi-instance attention.

\begin{table*}[t]\small
    \centering
    \setlength{\tabcolsep}{3pt}
    \caption{Quantitative comparisons on synthetic datasets~\cite{fu20213dfront,azinovic2022neuralrgbd} in scene-level Chamfer Distance (CD-S) and F-Score (F-Score-S), object-level Chamfer Distance (CD-O) and F-Score (F-Score-O), and Volume IoU of object bounding boxes (IoU-B).}
    \label{tab:comp_metrics}
    \begin{tabular}{l|ccccc|ccccc|c}
    \toprule
    \multirow{2}{*}{Method} & \multicolumn{5}{c|}{3D-Front} & \multicolumn{5}{c|}{BlendSwap} & \\
     & CD-S$\downarrow$ & F-Score-S$\uparrow$ & CD-O$\downarrow$ & F-Score-O$\uparrow$ & IoU-B$\uparrow$ & CD-S$\downarrow$ & F-Score-S$\uparrow$ & CD-O$\downarrow$ & F-Score-O$\uparrow$ & IoU-B$\uparrow$ & Runtime$\downarrow$ \\
    \midrule
    PanoRecon~\cite{dahnert2021panorecon} & 0.150 & 40.65 & 0.211 & 35.05 & 0.240 & 0.427 & 19.11 & 0.713 & 13.06 & 0.119 & 32s \\
    Total3D~\cite{nie2020total3d} & 0.270 & 32.90 & 0.179 & 36.38 & 0.238 & 0.258 & 37.93 & 0.168 & 38.14 & 0.328 & 39s \\
    InstPIFu~\cite{liu2022instpifu} & 0.138 & 39.99 & 0.165 & 38.11 & 0.299 & 0.129 & 50.28 & 0.167 & 38.42 & 0.340 & 32s \\
    SSR~\cite{chen2024ssr} & 0.140 & 39.76 & 0.170 & 37.79 & 0.311 & 0.132 & 48.72 & 0.173 & 38.11 & 0.336 & 32s \\
    DiffCAD~\cite{gao2024diffcad} & 0.117 & 43.58 & 0.190 & 37.45 & 0.392 & 0.110 & 52.83 & 0.169 & 38.98 & 0.457 & 64s \\
    Gen3DSR~\cite{dogaru2024gen3dsr} & 0.123 & 40.07 & 0.157 & 38.11 & 0.363 & 0.107 & 60.17 & 0.148 & 40.76 & 0.449 & 9min \\
    REPARO~\cite{han2024reparo} & 0.129 & 41.68 & 0.160 & 40.85 & 0.339 & 0.115 & 62.39 & 0.151 & 42.84 & 0.410 & 4min \\
    Ours & \textbf{0.080} & \textbf{50.19} & \textbf{0.103} & \textbf{53.58} & \textbf{0.518} & \textbf{0.077} & \textbf{78.21} & \textbf{0.090} & \textbf{62.94} & \textbf{0.663} & 40s \\
    \bottomrule
    \end{tabular}
\end{table*}

\section{Experiments}

\begin{figure*}
    \centering
    \includegraphics[width=\textwidth]{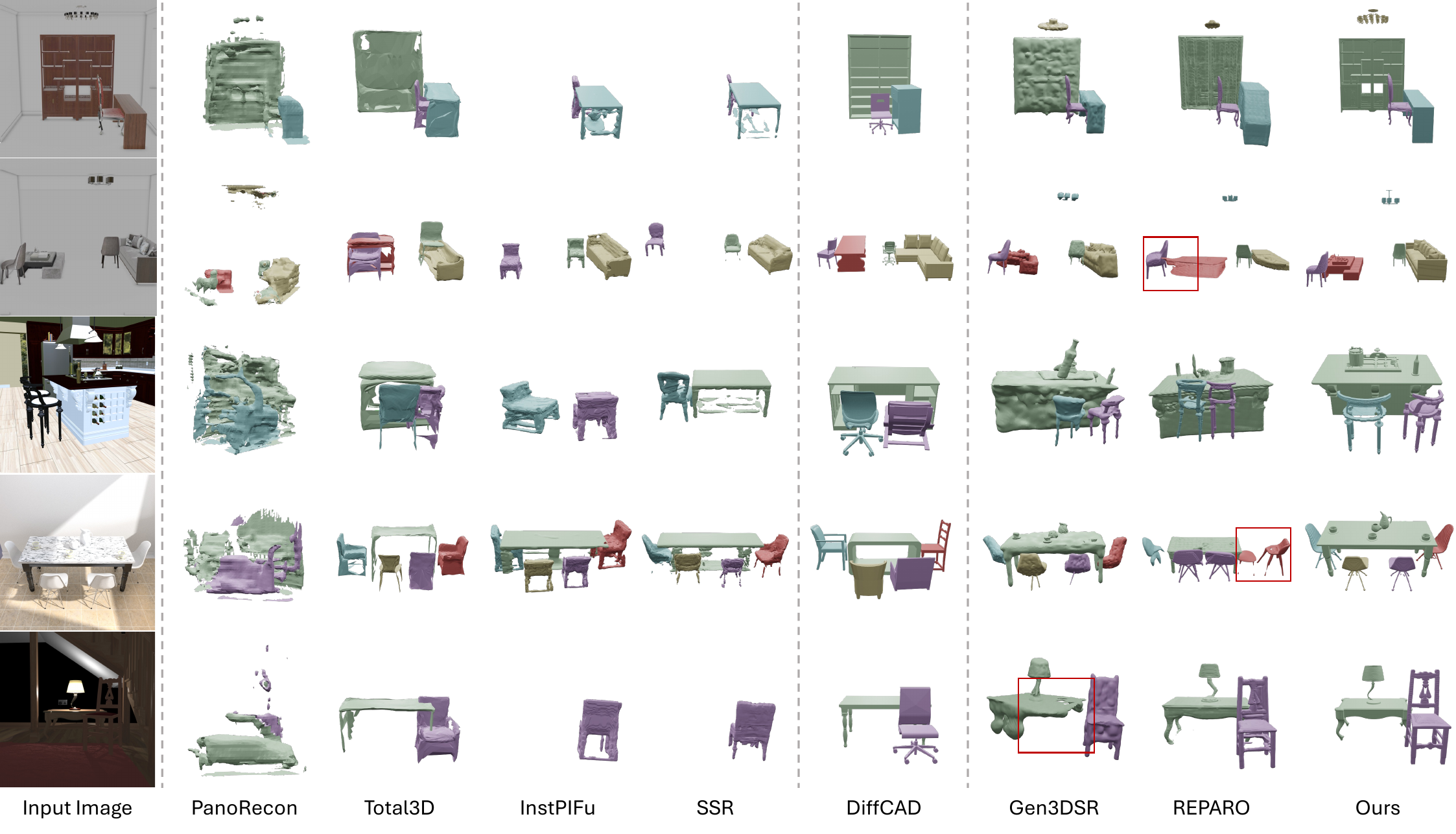}
    \caption{Qualitative comparisons on synthetic datasets, including 3D-Front~\cite{fu20213dfront} and BlendSwap~\cite{azinovic2022neuralrgbd}.}
    \label{fig:comparison_synthetic}
\end{figure*}

\begin{figure*}
    \centering
    \includegraphics[width=\textwidth]{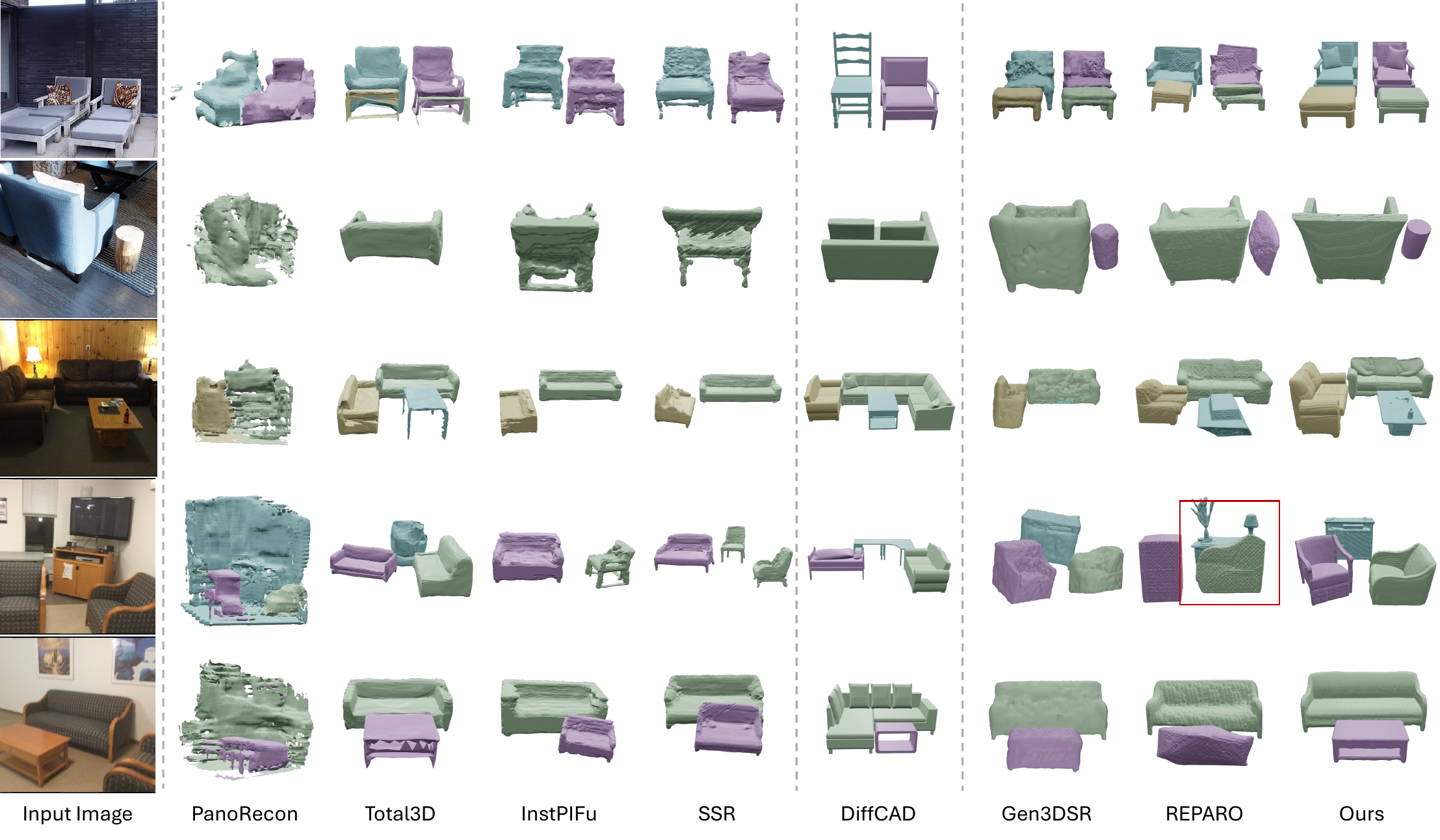}
    \caption{Qualitative comparisons on real-world data, including Matterport3D~\cite{chang2017matterport3d} and ScanNet~\cite{dai2017scannet}.}
    \label{fig:comparison_real}
\end{figure*}

\subsection{Setup}

\cparagraph{Implementation details.}
We implemented MIDI based on our own image-to-3D object generation model, which utilizes the rectified flow architecture~\cite{liu2022rflow} and employs 21 attention blocks to construct the denoising transformer network, developed from existing 3D object generation methods~\cite{zhao2024michelangelo,zhang2024clay}.
We initialize the image encoder $\tau_{\theta}$ in MIDI using DINO~\cite{oquab2023dinov2}, and expand the channel dimension of the input projection layer to accommodate 7 channels, corresponding to the concatenated inputs $\mathbf{y}$ (\ie scene images, object images and masks).
We set the resolution of $\mathbf{y}$ to 512.
During training, we adopt the Low-Rank Adaptation (LoRA) technique~\cite{hu2021lora} to fine-tune the pre-trained model efficiently.
For the multi-instance attention mechanism, we set the number of multi-instance attention layers $K$ to 5.
Note that we focus on generating the instances in the scene and their spatial relationships.
Planar background structures like floors and walls are not part of our generation scope, and they can be easily generated using existing methods~\cite{zhou2024deepprior,dogaru2024gen3dsr}.


\cparagraph{Datasets.}
We trained MIDI on the 3D-Front dataset~\cite{fu20213dfront}, which is a synthetic 3D dataset of indoor 3D scenes with rich annotations.
We performed cleaning by filtering out scenes with unreasonable object placements, such as intersecting or floating objects, resulting in approximately 15,000 high-quality scenes.
The dataset is split into training and testing sets, with 1,000 scene images randomly selected as the test set.
We evaluate MIDI on four widely used 3D scene reconstruction benchmarks, which includes synthetic datasets (\ie test set of 3D-Front~\cite{fu20213dfront}, BlendSwap~\cite{azinovic2022neuralrgbd}) and real-world datasets (\ie Matterport3D~\cite{chang2017matterport3d}, ScanNet~\cite{dai2017scannet}).
To further validate the generalization ability of MIDI, we also test on scene images with various styles generated by the text-to-image diffusion model~\cite{podell2023sdxl}.


\cparagraph{Baselines.}
We mainly compare our method with the state-of-the-art methods in scene reconstruction from single images, which includes feed-forward reconstruction methods PanoRecon~\cite{dahnert2021panorecon}, Total3D~\cite{nie2020total3d}, InstPIFu~\cite{liu2022instpifu} and SSR~\cite{chen2024ssr}, retrieval-based methods DiffCAD~\cite{gao2024diffcad}, and compositional generation methods Gen3DSR~\cite{dogaru2024gen3dsr} and REPARO~\cite{han2024reparo}.

\cparagraph{Metrics.}
Following existing scene reconstruction methods~\cite{nie2020total3d,zhou2024deepprior}, we use Chamfer Distance and F-Score with the default threshold of 0.1 to evaluate the whole scenes.
To further evaluate the geometric quality of individual 3D objects, we compute the Chamfer Distance and F-Score at the \emph{object level} for each object within the scene, assessing the fidelity of each object's geometry independently.
Additionally, we calculate Volumetric Intersection over Union (Volume IoU) between the bounding boxes of objects in the reconstructed or generated scene and those in the ground truth scene to assess the accuracy of object layouts and spatial arrangements.
We also report the average runtime for each method to generate one scene.

\subsection{Scene Generation on Synthetic Data}
\cref{tab:comp_metrics} reports quantitative comparisons on synthetic datasets, including 3D-Front~\cite{fu20213dfront} and BlendSwap~\cite{azinovic2022neuralrgbd}.
Our method, MIDI, achieves the best performance among the state-of-the-art methods across all evaluated metrics without incurring much time consumption.
Specifically, at the \textbf{object level}, our method significantly outperforms existing methods~\cite{dahnert2021panorecon,nie2020total3d,liu2022instpifu,chen2024ssr,gao2024diffcad,dogaru2024gen3dsr,han2024reparo} due to our novel design based on pre-trained 3D object prior.
Our MIDI, utilizing pre-trained object generation models, achieves a substantial leap in quality compared to methods that rely solely on reconstruction from limited data. 
At the \textbf{scene level}, metrics assessing the overall scene reconstruction quality and the alignment of object locations with ground truth demonstrate that our multi-instance diffusion models exhibit better robustness and accuracy compared to multi-stage object-by-object generation methods.
MIDI effectively models global scene knowledge and the spatial relationships between objects, resulting in coherent and accurately arranged scenes.

The qualitiative comparison is shown in \cref{fig:comparison_synthetic}.
Existing feed-forward reconstruction methods~\cite{dahnert2021panorecon,nie2020total3d,liu2022instpifu} often produce inaccurate geometry and misaligned scene layouts.
Retrieval-based methods~\cite{gao2024diffcad} produce results that do not accurately align with the input image.
Multi-stage object-by-object generation methods~\cite{dogaru2024gen3dsr,han2024reparo} generate instances that fail to align correctly with the overall scene due to the absence of scene context constraints during object image completion and 3D generation.
In contrast, MIDI produces high-quality geometries and preserves accurate spatial configurations among multiple instances, due to our utilization of pre-trained object priors and effective multi-instance attention mechanism.

\subsection{Scene Generation from Real Images}

We further evaluate MIDI on Matterport3D~\cite{chang2017matterport3d} and ScanNet~\cite{dai2017scannet} using real images.
For a qualitative comparison with other methods, we select 10 scenes from the test set of these two datasets, and sample one image from each scene as the input.
We show the visual comparisons in \cref{fig:comparison_real}, where we successfully generate scenes from real images and significantly outperform the previous works in the accuracy and completeness.
This demonstrates the huge potentials and generalization capabilities of the multi-instance diffusion models in generating real-world 3D scenes.

\begin{figure*}
    \centering
    \includegraphics[width=\textwidth]{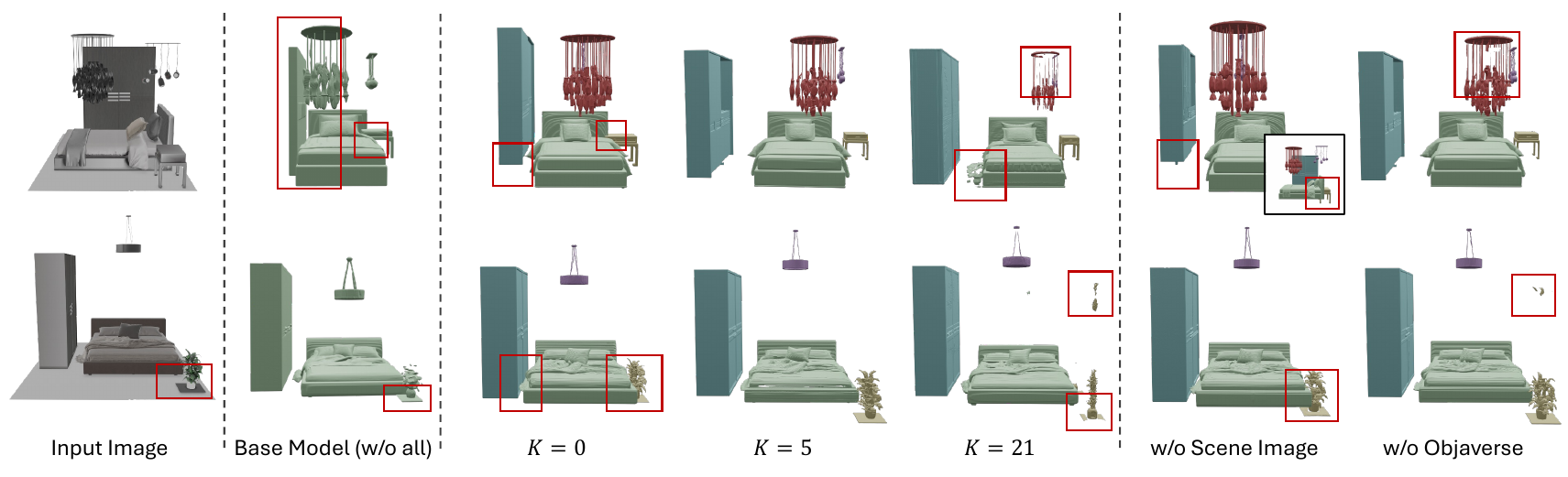}
    \caption{Qualitative ablation studies on the number of multi-instance attention layers $K$, and the use of global scene image conditioning, and mixed training with single-object dataset.}
    \label{fig:ablation}
\end{figure*}

\subsection{Scene Generation from Stylized Images}

\begin{figure}
    \centering
    \includegraphics[width=\linewidth]{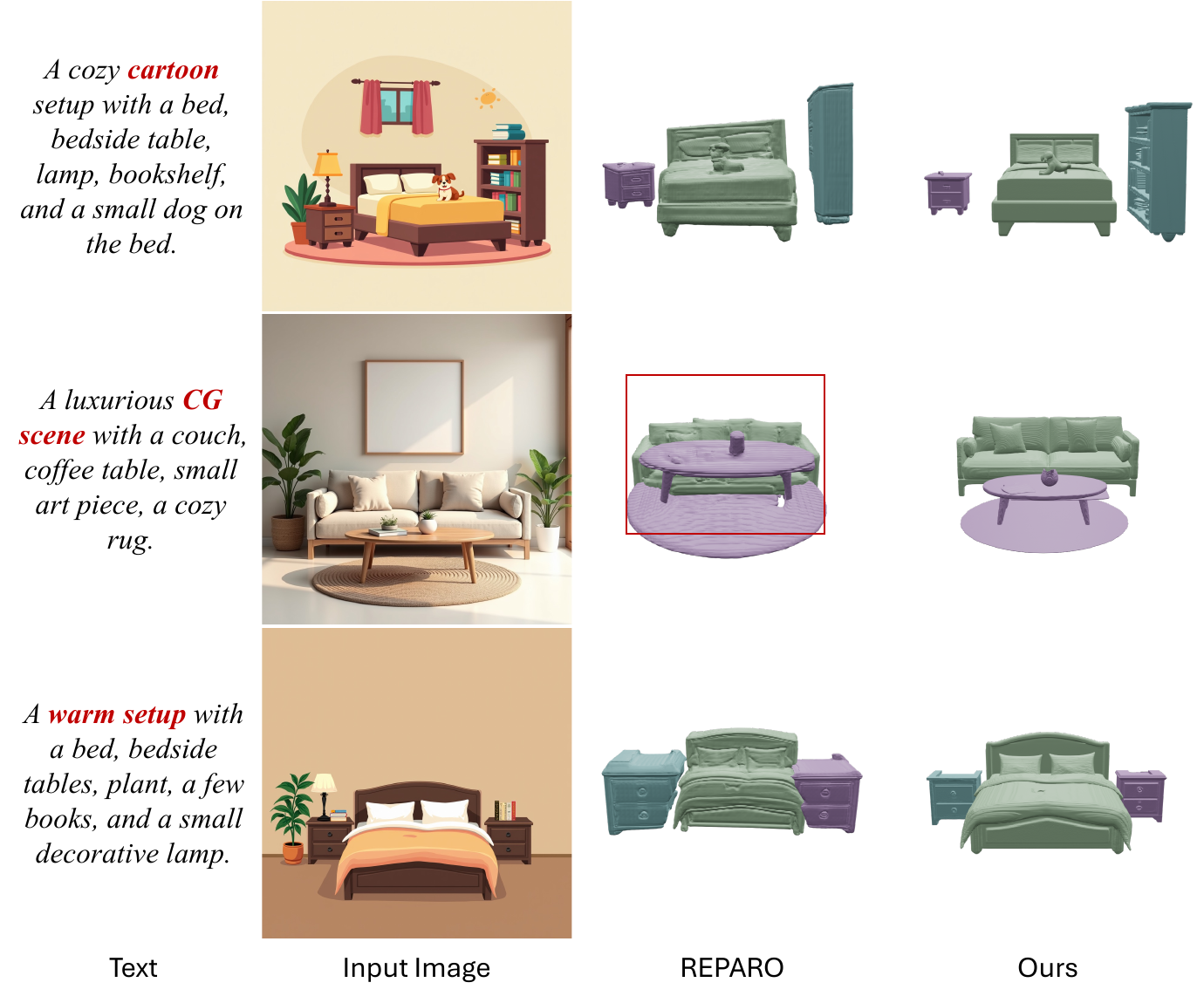}
    \caption{Qualitative comparisons on stylized images that are generated by text-to-image diffusion models.}
    \label{fig:comparison_t2i}
\end{figure}

To further assess the generalization capabilities of MIDI, we utilize the text-to-image diffusion model SDXL~\cite{podell2023sdxl} to generate scene images with diverse styles, and test our method on them.
Due to the limitations of existing methods in handling such diverse inputs, we compare MIDI exclusively with REPARO~\cite{han2024reparo}.
As shown in \cref{fig:comparison_t2i}, MIDI generates accurate and coherent 3D scenes from the varied input images, demonstrating its strong generalization ability.

\subsection{Ablation Study}

\begin{table}\footnotesize
\caption{Ablation studies. We evaluate the number of multi-instance attention layers ($\#K$), the inclusion of global scene image (S.) input, and the use of Objaverse~\cite{deitke2023objaverse} (O.) for mixed training.}
\label{tab:ablation}
    \centering
    \setlength{\tabcolsep}{3pt}
    \begin{tabular}{c|c|c|ccccc}
   \toprule
   $\#K$ & S. & O. & CD-S$\downarrow$ & F-Score-S$\uparrow$ & CD-O$\downarrow$ & F-Score-O$\uparrow$ & IoU-B$\uparrow$ \\
   \midrule
   $0$ & \XSolidBrush & \XSolidBrush & 0.152 & 41.16 & -- & -- & -- \\
   \midrule
   $0$ & \Checkmark & \Checkmark & 0.145 & 40.94 & 0.096 & 54.16 & 0.327 \\
   $5$ & \Checkmark & \Checkmark & \textbf{0.080} & \textbf{50.19} & \textbf{0.103} & \textbf{53.58} & \textbf{0.518} \\
   $21$ & \Checkmark & \Checkmark & 0.127 & 44.88 & 0.141 & 48.55 & 0.423 \\
   \midrule
   $5$ & \XSolidBrush & \Checkmark & 0.134 & 41.49 & 0.102 & 52.91 & 0.459 \\
   $5$ & \Checkmark & \XSolidBrush & 0.137 & 42.00 & 0.126 & 51.62 & 0.502 \\
   \bottomrule
\end{tabular}
\end{table}

We conduct ablation studies on 3D-Front~\cite{fu20213dfront} dataset to evaluate the impact of key components in MIDI.
Specifically, we examine:
1) the number of multi-instance attention layers $K$,
2) the inclusion of the global scene image as conditioning input,
and 3) the use of single-object dataset~\cite{deitke2023objaverse} for mixed training.

\cparagraph{Base model without any design.}
We start with a baseline that directly fine-tunes the object generation model on the scene dataset without any design.
However, the baseline model can not generate separable multi-instances, and shows weak modeling of spatial relationships (see \cref{fig:ablation}) due to limited scene data for training.

\cparagraph{Number of multi-instance attention layers $K$.}
We experiment with $K=0$, $K=5$, and $K=21$.
Quantitative results in \cref{tab:ablation} and qualitative examples in \cref{fig:ablation} indicate that $K=5$ achieves the best performance.
With $K=0$, the model fails to capture correct spatial relationships, leading to incoherent scene layouts, demonstrating the importance of our proposed multi-instance attention.
When $K=21$, excessive attention layers cause overfitting and distorted object geometries due to disruption of the pre-trained 3D prior after the model is trained on a relatively small scene dataset.
We choose $K=5$, where only a subset of the self-attention layers are converted to multi-instance attention, balancing between modeling interactions and preserving the pre-trained prior.

\cparagraph{Global scene image conditioning.}
We remove the global scene image from the input and condition the model solely on local object images and masks.
As shown in \cref{tab:ablation} and \cref{fig:ablation}, excluding the global scene context significantly impairs the model's ability to generate coherent 3D scenes.
The resulting scenes exhibit incorrect object placements and lack proper spatial relationships among instances.

\cparagraph{Mixed training with single-object dataset.}
We explore the effect of mixed training by incorporating the Objaverse dataset~\cite{deitke2023objaverse} into the training process.
Results in \cref{tab:ablation} and \cref{fig:ablation} show that, without this regularization, the model tends to produce objects with inferior geometry, as it overfits on the smaller scene dataset.
Including single-object data helps preserve the object-level knowledge, enabling the model to generate high-quality geometries while effectively modeling inter-object interactions.

\section{Conclusion}

\cparagraph{Limitations and future works.}
MIDI performs relatively poorly for tiny-resolution image input and complex inter-instance interaction, as shown in the supplementary materials.
Building upon our proposed multi-instance diffusion for compositional 3D scene generation, future work can explore several directions:
1) extending the approach to model more complex interactions in compositional scenes, such as characters interacting with objects (\eg ``a panda playing a guitar''), which requires specialized datasets;
2) incorporating explicit 3D geometric knowledge to develop more efficient and expressive multi-instance attention mechanisms;
3) investigating the latent, implicit 3D perception capabilities of scene generation models;
and 4) scaling the framework to handle a larger number of objects and operate in open-world environments.


\cparagraph{Conclusion.}
This paper introduces MIDI, an innovative approach that significantly advances 3D scene generation from a single image.
By extending pre-trained image-to-3D object generation models to multi-instance diffusion models and incorporating a novel multi-instance attention mechanism, MIDI effectively captures complex inter-object interactions and spatial coherence directly within the generation process.
This enables the simultaneous generation of multiple 3D instances with accurate spatial relationships, leading to high-quality 3D scenes with precise geometries and spatial layouts.
Extensive experiments demonstrate that MIDI achieves state-of-the-art performance while exhibiting strong generalization capabilities.

\section*{Acknowledgment}

This work was supported by National Natural Science Foundation of China (62132001), and the Fundamental Research Funds for the Central Universities.

{
    \small
    \bibliographystyle{ieeenat_fullname}
    \bibliography{main}
}

\clearpage
\setcounter{page}{1}

\maketitlesupplementary

\begin{figure*}
    \centering
    \includegraphics[width=\textwidth]{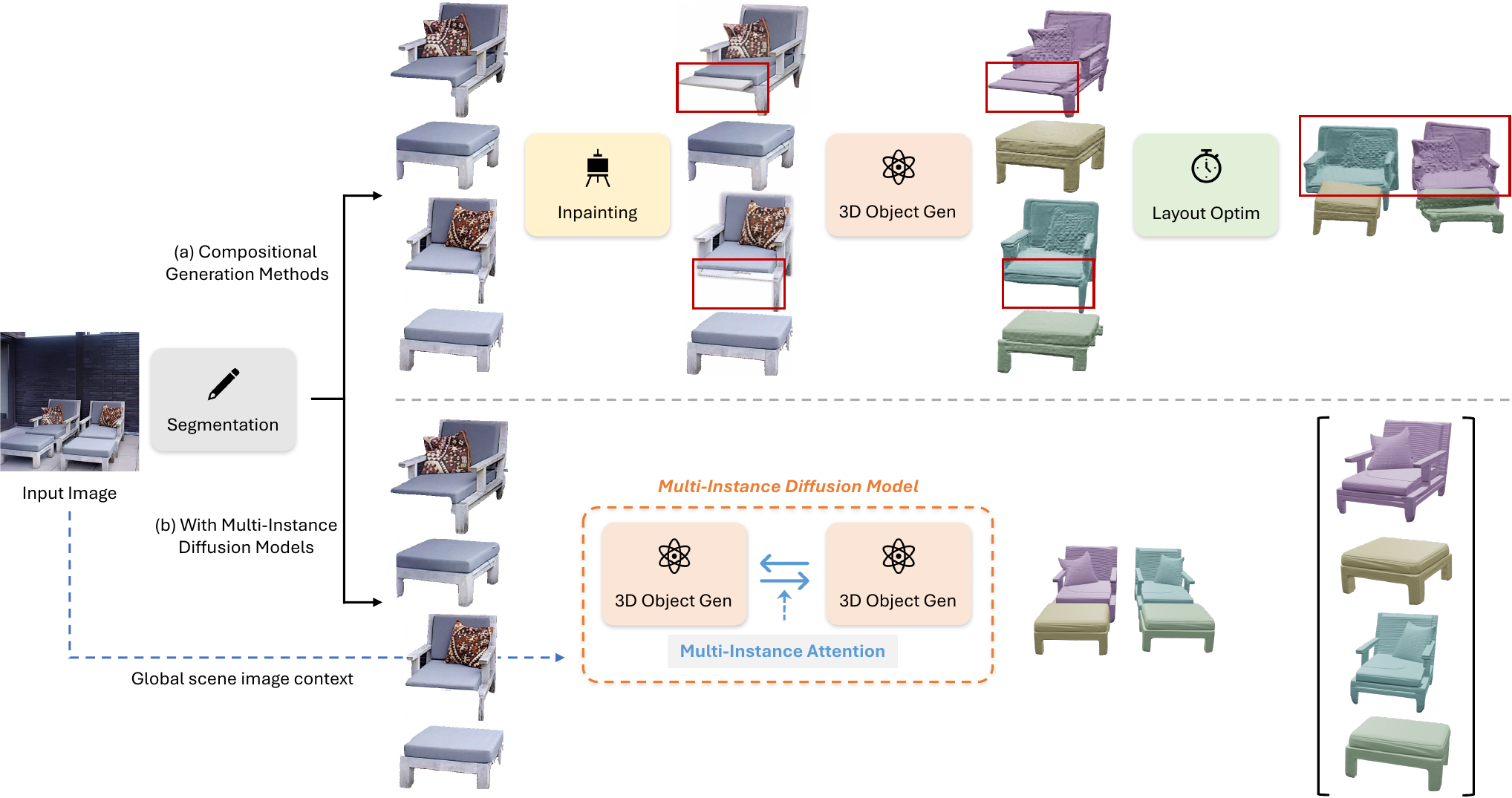}
    \caption{Detailed comparison between existing compositional generation methods and our multi-instance diffusion.}
    \label{fig:pipeline_diff_detail}
\end{figure*}

\section{Background}
\label{sec:background}

\cparagraph{Base model.}
Following scalable 3D object generation methods~\cite{zhao2024michelangelo,zhang2024clay,wu2024direct3d,li2024craftsman}, we firstly trains a VAE to compress 3D geometric representations into a low-dimensional latent space.
Specifically, $\mathbf{x}\in \mathbb{R}^{L\times 6}$, which represents positions and normals of $L$ points, are mapped to latent space by $\mathbf{z}=\mathcal{E}(\mathbf{x})$, where $\mathbf{z}\in \mathbb{R}^{l\times c}$, and $l$ denotes the length of the tokens after compression.
The latents are converted back to the 3D space by regressing signed distance function (SDF) values using $\mathbf{s}=\mathcal{D}(\mathbf{z})$.
Following 3DShape2Vecset~\cite{zhang20233dshape2vecset}, the VAE comprises of several transformer blocks.

Next, the denoising network $\epsilon_{\theta}$ is trained in the compressed latent space to transform noise $\bm{\epsilon}\sim\mathcal{N}(0,I)$ into the original 3D data distribution.
During training, following the rectified flow architecture~\cite{liu2022rflow}, the original data $\mathbf{z}_{0}$ is perturbed along a simple linear trajectory:
\begin{equation}
    \mathbf{z}_{t} = t \mathbf{z}_{0} + (1-t) \bm{\epsilon}
\end{equation}
for $t=1,\cdots,T$, where $T$ represents the number of steps in the diffusion process.
In practice, we adopt logit-normal sampling~\cite{esser2024sd3} to increase the weight for intermediate steps.
The denoising network $\epsilon_{\theta}$, featuring 21 attention blocks with residual connections, is trained to approximate the slope of the distribution transformation trajectory by minimizing the following loss:
\begin{equation}
    \mathbb{E}_{\mathbf{z}, \mathbf{y}, \bm{\epsilon} \sim\mathcal{N}(0,I), t} [\lVert \mathbf{z}_{0} - \bm{\epsilon} - \epsilon_{\theta}(\mathbf{z}_{t}, t, \tau_{\theta}(\mathbf{y}) ) \rVert_{2}^{2}]
\end{equation}
where $\tau_{\theta}$ is the image encoder, and $\mathbf{y}$ is the conditioning image, incorporated into the denoising transformer via cross-attention mechanism.

\section{Implementation Details}

\cparagraph{Training.}
we trained MIDI to simultaneously generate up to $N=7$ instances.
We selected this value based on an analysis of the 3D-FRONT dataset~\cite{fu20213dfront}, where we observed that scenes containing five or fewer objects constitute the majority, while scenes with more than five objects are relatively rare.
Instead of excluding scenes with more than $5$ objects, we employed a clustering method to select five representative objects from such scenes for training.
During training, we randomly dropped the image conditioning with a probability of 0.1.
We adopted the same strategy as in the training of the base model, utilizing logit-normal sampling~\cite{esser2024sd3} to increase the weight of intermediate diffusion steps, which helps the model focus on the more challenging stages of the generation process.
For the training configuration, we used a learning rate of $5\times 10^{-5}$ and trained MIDI for 5 epochs on 8 NVIDIA A100 GPUs.

\cparagraph{Inference.}
In our experimental setup, we first used Grounded-SAM~\cite{ren2024groundedsam} to segment the scene images, obtaining masks for individual objects.
We then applied our multi-instance diffusion model to generate compositional 3D instances using classifier-free guidance~\cite{ho2022cfg}, which enhances the fidelity and coherence of the generated scenes.
We set the number of inference steps to 50 and the guidance scale to 7.0.
The entire process of generating a 3D scene from a single image takes approximately 40 seconds on an NVIDIA A100 GPU.

\begin{table}\small
\caption{Training costs. (Batch size is set to 1)}
\label{tab:training_cost}
    \centering
    \begin{tabular}{l|cc}
   \toprule
   Number of Instances $N$ & VRAM (GB) & Speed (iter/s) \\
   \midrule
   $N=1$ & 15 & 1.50 \\
   $N=3$ & 17 & 0.83 \\
   $N=5$ & 19 & 0.55 \\
   $N=7$ & 21 & 0.40 \\
   \bottomrule
\end{tabular}
\end{table}

\section{Additional Discussions}

\cparagraph{MIDI vs. compositional generation methods.}
As show in \cref{fig:pipeline_diff_detail}, existing compositional generation methods involve a multi-step process, generating 3D objects one by one and then optimizing their spatial relationships.
However, this type of methods lack the contextual information of the global scene when generating objects, thus generating inaccurate or mismatched 3D objects.
In addition, it is very difficult to optimize the accurate scene layout based on a single image, and the position of similar objects will be reversed when there are similar objects in the scene (as shown in \cref{fig:pipeline_diff_detail}).
In contrast, our method models object completion, 3D generation and spatial relationships in a multi-instance diffusion model, thus generating coherent and accurate 3D scenes.

\cparagraph{Training costs.}
Table \ref{tab:training_cost} presents the training costs for MIDI.
As the number of instances $N$ increases, both GPU memory requirements and training time increase.
However, even when $N=7$, resource utilization remains manageable, demonstrating the scalability of MIDI.

\begin{figure}
    \centering
    \includegraphics[width=\linewidth]{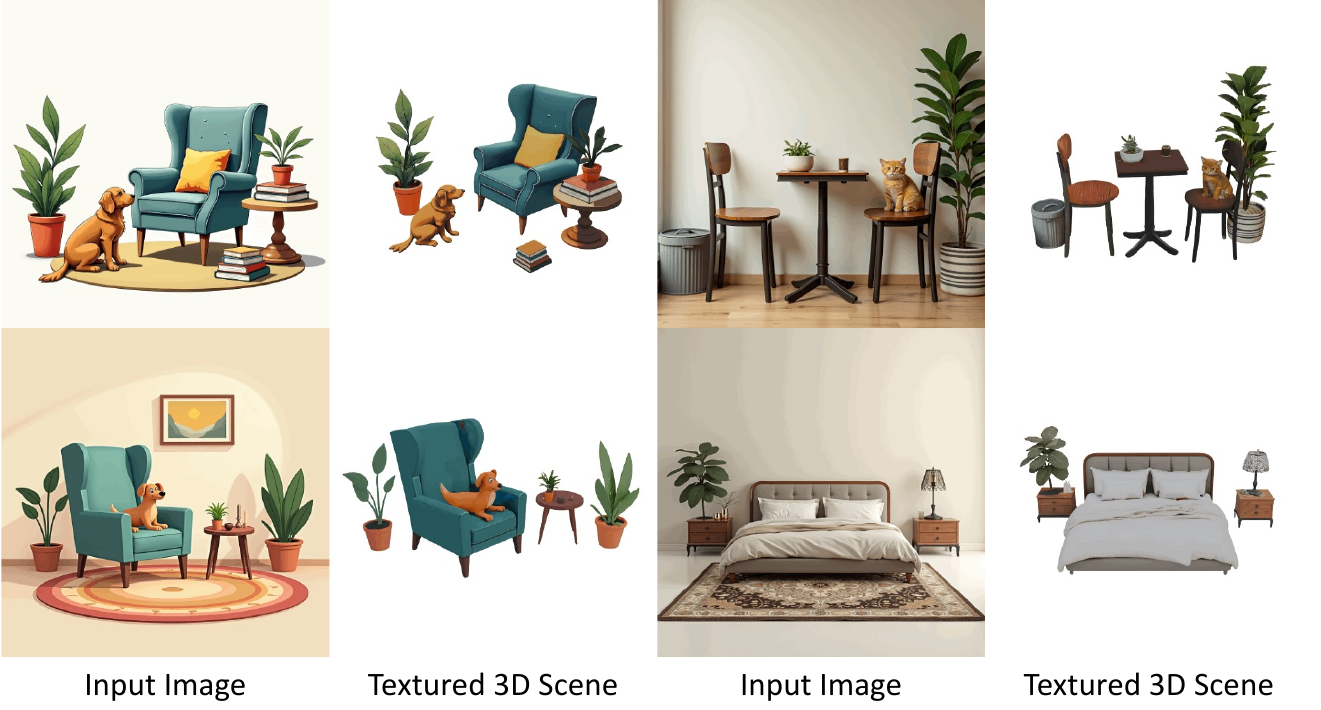}
    \caption{Visualization results of textured 3D scene generation with MV-Adapter~\cite{huang2024mvadapter}.}
    \label{fig:texture_gen}
\end{figure}

\cparagraph{Texture generation.}
To generate textured 3D scene from single images, we firstly synthesize 3D geometry with our MIDI, and then leverage MV-Adapter~\cite{huang2024mvadapter} to generate texture for each instance with the partial image of instance image as input.
The visualization results are shown in \cref{fig:texture_gen}.
It is recommended to interactively experience the generated 3D scenes in \href{https://huanngzh.github.io/MIDI-Page/}{our project page}.

\begin{figure}
    \centering
    \includegraphics[width=\linewidth]{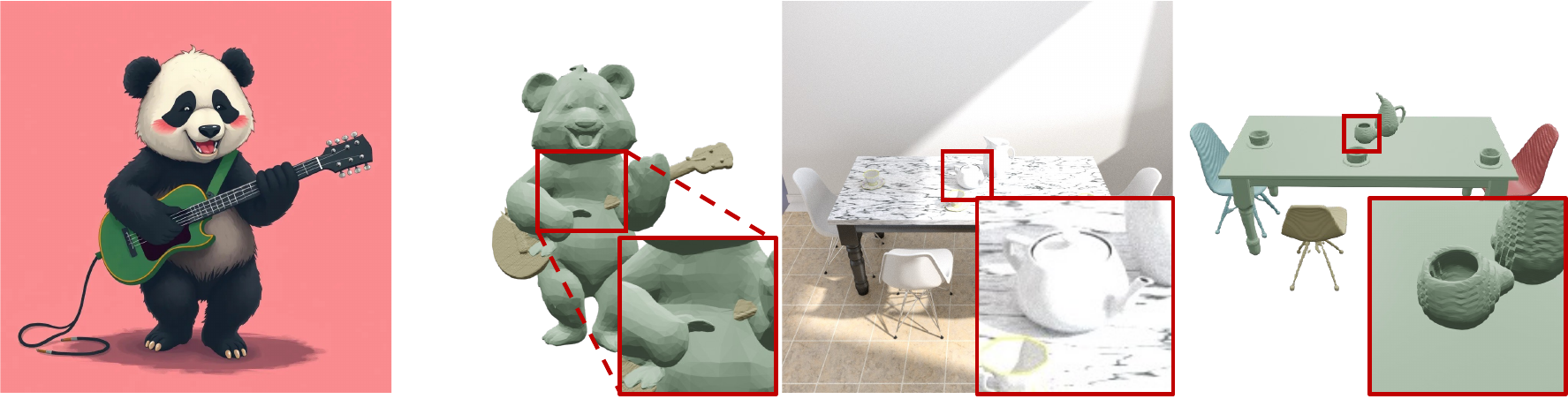}
    \caption{Failure cases.}
    \label{fig:error}
\end{figure}

\section{Limitations}

We present two typical failure examples of MIDI in \cref{fig:error}.
While MIDI generates 3D instances within the global scene coordinate system—specifically, a normalized space ranging from $-1$ to $1$—this approach causes smaller objects to occupy a relatively minor portion of the overall space.
Consequently, these small objects may have lower resolution compared to objects generated in their canonical spaces, where the entire capacity of the model can focus on a single object.
We believe that enhancing the multi-instance diffusion model to generate objects in their canonical spaces, along with their spatial positions within the scene, could address this issue by allowing each object to be generated at optimal resolution.

Also, our model is constrained by the simplicity of interaction relationships present in existing scene datasets.
As a result, MIDI may struggle to generate scenes featuring intricate interactions, such as objects with dynamic interplays.
We anticipate that introducing more complex and diverse training data, encompassing a wider variety of object interactions and spatial relationships, would enhance the model's capacity to generalize at the level of object spatial interactions.
This improvement would enable the generation of scenes with more sophisticated and realistic inter-object dynamics.


\end{document}